\documentclass[a4paper,twoside]{article}

\usepackage{epsfig}
\usepackage{subfigure}
\usepackage{calc}
\usepackage{amssymb}
\usepackage{amstext}
\usepackage{amsmath}
\usepackage{amsthm}
\usepackage{multicol}
\usepackage{multirow}
\usepackage{pslatex}
\usepackage{apalike}
\usepackage{SCITEPRESS}
\usepackage[small]{caption}
\usepackage{array}

\begin{document}

\newcolumntype{L}[1]{>{\RaggedRight\hspace{0pt}}p{#1}}
\newcolumntype{R}[1]{>{\RaggedLeft\hspace{0pt}}p{#1}}

\title{A Pattern Recognition System for Detecting Use of Mobile Phones While Driving}

\author{\authorname{Rafael A. Berri\sup{1}, Alexandre G. Silva\sup{1}, Rafael S. Parpinelli\sup{1}, Elaine Girardi\sup{1} and Rangel Arthur\sup{2}}
\affiliation{\sup{1}College of Techological Science, Santa Catarina State University (UDESC), Joinville, Brazil}
\affiliation{\sup{2}Faculty of Technology, University of Campinas (Unicamp), Limeira, Brazil}
\email{rafaelberri@gmail.com, \{alexandre, parpinelli\}@joinville.udesc.br, elaine\_girardi@hotmail.com, rangel@ft.unicamp.br}
}


\abstract{It is estimated that 80\% of crashes and 65\% of near collisions involved drivers inattentive to traffic for three se\-conds before the event. This paper develops an algorithm for extracting characteristics allowing the cell phones identification used during driving a vehicle. Experiments were performed on sets of images with 100 positive images (with phone) and the other 100 negative images (no phone), containing frontal images of the driver. Support Vector Machine (SVM) with Polynomial kernel is the most advantageous classification system to the features provided by the algorithm, obtaining a success rate of 91.57\% for the vision system. Tests done on videos show that it is possible to use the image datasets for training classifiers in real situations. Periods of 3 seconds were correctly classified at 87.43\% of cases.}

\onecolumn \maketitle \normalsize \vfill

\section{\uppercase{Introduction}}
\label{sec:intro}

The distraction whiles driving \cite{regan2008,peissner2011}, ie, an action that diverts the driver's attention from the road for a few seconds, re\-pre\-sents about half of all cases of traffic accidents. Dialing a telephone number, for example, consumes about $5$ seconds, resulting in $140$ meters tra\-ve\-led by an automobile at $100$ km/h \cite{balbinot2011}. In a study done in Washington by Virginia Tech Transportation Institute revealed, after $43,000$ hours of testing, that almost $80\%$ of crashes and $65\%$ of near collisions involved drivers who were not paying enough attention to traffic for three seconds before the event.

About $85\%$ of American drivers use cell phone while driving \cite{goodman1997}. 
At any daylight hour $5\%$ of cars on U. S. roadways are driven by people on phone calls \cite{nhtsa2011}.
Driver distraction has the three main causes: visual (eyes off the road), ma\-nu\-al (hands off the wheel), and cognitive (mind off the task) \cite{strayer2011}.
Talking on a hand-held or handsfree cell phone increases substantially the cognitive distraction \cite{strayer2013}.


This work proposes a pattern recognition system to detect hand-held cell phone use du\-ring the act of driving to try to reduce these numbers. The main goal is to present an algorithm for the characteristics extraction which identifies the cell phone use, producing a warning that can regain the driver's attention exclusively to the vehicle and the road. The aim is also to test classifiers and choose the technique that maximizes the accuracy of the system. In Section~\ref{sec:trabrel} related works are described. Support tools for classification are presented in Section~\ref{sec:definicoes}. In Section~\ref{sec:extracao} the algorithm for feature extraction is developed. In Section~\ref{sec:experimentos} experiments performed on an image database are shown. And finally conclusions are set out in Section~\ref{sec:conclusao}.

\section{\uppercase{Related works}}
\label{sec:trabrel}

The work produced by \cite{veeraraghavan2007} is the closest to this paper. The authors' goal is to make detection and classification of ac\-ti\-vi\-ties of a car's driver using an algorithm that detects relative movement and the segmentation of the skin of the driver. Therefore, its operation depends on obtaining a set of frames, and the need to put a camera sideways (in relation to the driver) in the car, impeding the presence of passengers beside the driver.

The approach of \cite{yang2011} use custom beeps of high frequency sent through the car sound equipment, network \textit{Bluetooth}, and a software running on the phone for capturing and processing sound signals. The purpose of the beeps are to estimate the position in which the cell phone is, and to dif\-fe\-ren\-tia\-te whether the driver or another passenger in the car is using it. The proposal obtained accuracy classification of more than 90\%. However, the system depends on the operating system and mobile phone brand, and the software has to be continually e\-na\-bled by the driver. On the other hand, the approach works even if there is use of headphones (hands-free).

The study of \cite{enriquez2009} segments the skin, a\-na\-ly\-zing two color spaces, \textit{YCrCb} and \textit{LUX}, and using specifically the coordinates \textit{Cr} and \textit{U}. The advantage is the ability to use a simple camera (webcam) for image acquisition.

Another proposal by \cite{watkins2011} is autonomously to identify distracted driver behaviors associated with text messaging on devices. The approach uses a cell phone programmed to record any typing done. An analysis can be performed to verify distractions through these records.

\section{\uppercase{Preliminary definitions}}
\label{sec:definicoes}

In this section, support tool for classifying classes are presented.

\subsection{Support Vector Machines (SVM)}
\label{sec:svm}

The SVM (Support Vector Machine) was introduced by Vapnik in 1995 and is a tool for binary classification \cite{vapnik1995}. Given data set $\lbrace (\vec{x}_1,y_1), \cdots, (\vec{x}_n,y_n) \rbrace$ with input data $\vec{x}_i \in R^d$ (where $d$ is the dimensional space) and output $y$ labeled as $y \in \lbrace-1, +1\rbrace$. The central idea of the technique is to ge\-ne\-ra\-te an optimal hyperplane chosen to maximize the separation between the two classes, based on support vector \cite{wang2005}. The training phase consists in the choice of support vectors using the training data before labeled. 



From SVM is possible to use some \textit{kernel} functions for the treating of nonlinear data. The kernel function transforms the original data into a space of features of high dimensionality, where the nonlinear relationships can be present as linear \cite{stanimirova2010}. Among the existing kernels, there are
Linear (Equation~\ref{eq:svmlinear}),
Polynomial (Equation~\ref{eq:svmpoli}), 
Radial basis (Equation~\ref{eq:svmrbf}), and 
Sigmoid (Equation~\ref{eq:svmsigmoide}).
The choice of a suitable function and correct parameters are an important step for achieving the high accuracy of the classification system.

\begin{equation}
  \label{eq:svmlinear}
  K(\vec{x}_i, \vec{x}_j) = \vec{x}_i \cdot \vec{x}_j
\end{equation} 
\begin{equation}
  \label{eq:svmpoli}
  K(\vec{x}_i, \vec{x}_j) = (\gamma(\vec{x}_i \cdot \vec{x}_j) + coef0)^{degree}, \rm{where} \; \gamma > 0
\end{equation} 
\begin{equation}
  \label{eq:svmrbf}
  K(\vec{x}_i, \vec{x}_j) = e^{-\gamma \parallel (\vec{x}_i + \vec{x}_j) \parallel ^2}, \rm{where} \; \gamma > 0
\end{equation} 
\begin{equation}
  \label{eq:svmsigmoide}
  K(\vec{x}_i, \vec{x}_j) = tanh(\gamma (\vec{x}_i \cdot \vec{x}_j) + coef0)
\end{equation} 


We can start the training after the choice of the \textit{kernel} function. We need to maximize the values for $\vec{\alpha}$ in Equation~\ref{eq:vetor}. This is a quadratic programming problem \cite{hearst1998} and it is subject to the constraints (for any $i = 1, ..., n$ where $n$ is the amount of training data): $0 \leq \alpha_i \leq C$ e $\sum_{i = 1}^{n} \alpha_i y_i = 0$. The penalty parameter $C$ has the ratio of the algorithm complexity and the number of wrongly classified training samples.

\begin{equation}
  \label{eq:vetor}
  W( \vec{\alpha} ) = \sum_{i = 1}^{n} \alpha_i - \frac{1}{2} \sum_{i,j=1}^{n} \alpha_i \alpha_j y_i y_j K(\vec{x}_i, \vec{x}_j)
\end{equation} 


The threshold $b$ is found with Equation~\ref{eq:svmlimiar}. The calculation is done for all the support vectors $\vec{x}_j$ ($0 \leq \alpha_j \leq C$). The $b$ value is equal to the average of all calculation.

\begin{equation}
  \label{eq:svmlimiar}
  b = y_j - \sum_{i = 1}^{l} y_i \alpha_i K(\vec{x}_i, \vec{x}_j)
\end{equation} 


A feature vector $\vec{x}$ can be classified (prediction) with Equation~\ref{eq:svmclassifica}, where $\lambda_i = y_i \alpha_i$ and mathematical function sign extracts the sign of a real number (returns: $-1$ for negative, $0$ for zero value and $+1$ for positive values​​). 

\begin{equation}
  \label{eq:svmclassifica}
  f(\vec{x}) = sign(\sum_{i}^{} \lambda_i K(\vec{x}, \vec{x}_i) + b)
\end{equation} 


This SVM\footnote{The library LibSVM is used (available in {\scriptsize \texttt{http://www.csie.ntu.edu.tw/~cjlin/libsvm}}).} uses imperfect separation, and the $nu$ is used as a replacement for $C$. The parameter $nu$ uses values between $0$ and $1$. If this value is higher, the decision boundary is smoother.

\section{\uppercase{Extraction of features}}
\label{sec:extracao}

Driving act makes drivers instinctively and continually look around. But with the use of the cell phone, the tendency is to fix the gaze on a point in front, effecting the drivability and the attention in the traffic by limiting the field of vision \cite{drews2008}. Starting from this principle, this work choses an algorithm that can detect the use of cell phones by drivers just by using the frontal camera attached on the dashboard of a car. The following subsections explain the details.

\subsection{Algorithm}
\label{sec:algoritmo}



In general, the algorithm is divided into the following parts:
\begin{description}
  \item [Acquisition:]
        Driver image capture system.
  \item[Preprocessing:]
        Location of the driver, cropping the interest region (Section~\ref{sec:prepro}).
  \item[Segmentation:]
        Isolate the driver skin pixels (Section~\ref{sec:segment}).
  \item[Extraction of features:]
        Extraction of skin per\-cen\-ta\-ge in regions where usually the driver's hand/arm would be when using the cell phone, and calculation of HU's Moments \cite{hu1962} (Section~\ref{sec:carac}).
\end{description}

In the following subsections, the steps of the algorithm (after the image acquisition) are explained.

\subsubsection{Preprocessing}
\label{sec:prepro}

After image acquisition, the preprocessing step is tasked to find the region of interest, ie, the face of the driver. In this way, three detectors\footnote{Haar-like-features from OpenCV ({\scriptsize\texttt{http://opencv.org}}) with three training files: haarcascade\_frontalface\_alt2.xml, haarcascade\_profileface.xml and haarcascade\_frontalface\_default.xml} are applied based on \textit{Haar-like-features} for feature extraction and \textit{Adaboost} as classifier \cite{viola2001}. The algorithm adopts the largest area found by these detectors as being the driver’s face.


The region found is increased by 40\% in width, 20\% to the left and 
20\% to the right of the face because often this is much reduced (allows only the face to be visible), becoming impossible to detect the hand/arm. In Figures~\ref{fig:faceregcom} and \ref{fig:faceregsem}, preprocessing results are exemplified.


\subsubsection{Segmentation}
\label{sec:segment}

The segmentation is based on locating pixels of the skin of the driver present in the preprocessed i\-ma\-ge. It is necessary, as a first step, converting the image into two different color spaces: \textit{Hue Saturation Value} (HSV) \cite{smith1996} and \textit{Luminance--chrominance} YCrCb \cite{dadgostar2006}.

After the initial conversion, a reduction of the complexity of the three components from YCbCr and HSV is applied. Each component has $256$ possible levels ($8$ bits) resulting in more than $16$ million possibilities ($256^3$) by color space. Segmentation is simplified by reducing the total number of possibilities in each component. The level of each component is divided by 32. Thus, each pixel has 512 possibilities $((\frac{256}{32})^3)$ of representation.

A sample of skin is then isolated with size of 10\% in the height and 40\% in width of the image, being centralized in the width and with the top in center of the height, according to the rectangles in Figures~\ref{fig:faceregcom} and \ref{fig:faceregsem}. 
A histogram with the 3 dimensions of HSV and YCrCb skin sample from this region is crea\-ted by counting the pixels that have the same levels in all 3 components. 
Further, a threshold is performed in the image with the levels that represent at least 5\% of the pixels contained in the sample region.
Though the skin pixels for HSV and YCrCb are obtained se\-pa\-ra\-te\-ly, the results are the pixels that are in both segmentation of HSV and YCrCb.
In Figures~\ref{fig:regcom} and \ref{fig:regsem}, segmentation results are exemplified.


\subsubsection{Features}
\label{sec:carac}

Two features are part of the driver's classification with or without cell phone: Percentage of the Hand ($PH$) and Moment of Inertia ($MI$).

\begin{figure}[ht]
  \center
  \subfigure[Crop (with phone)]{\includegraphics[width=60px]{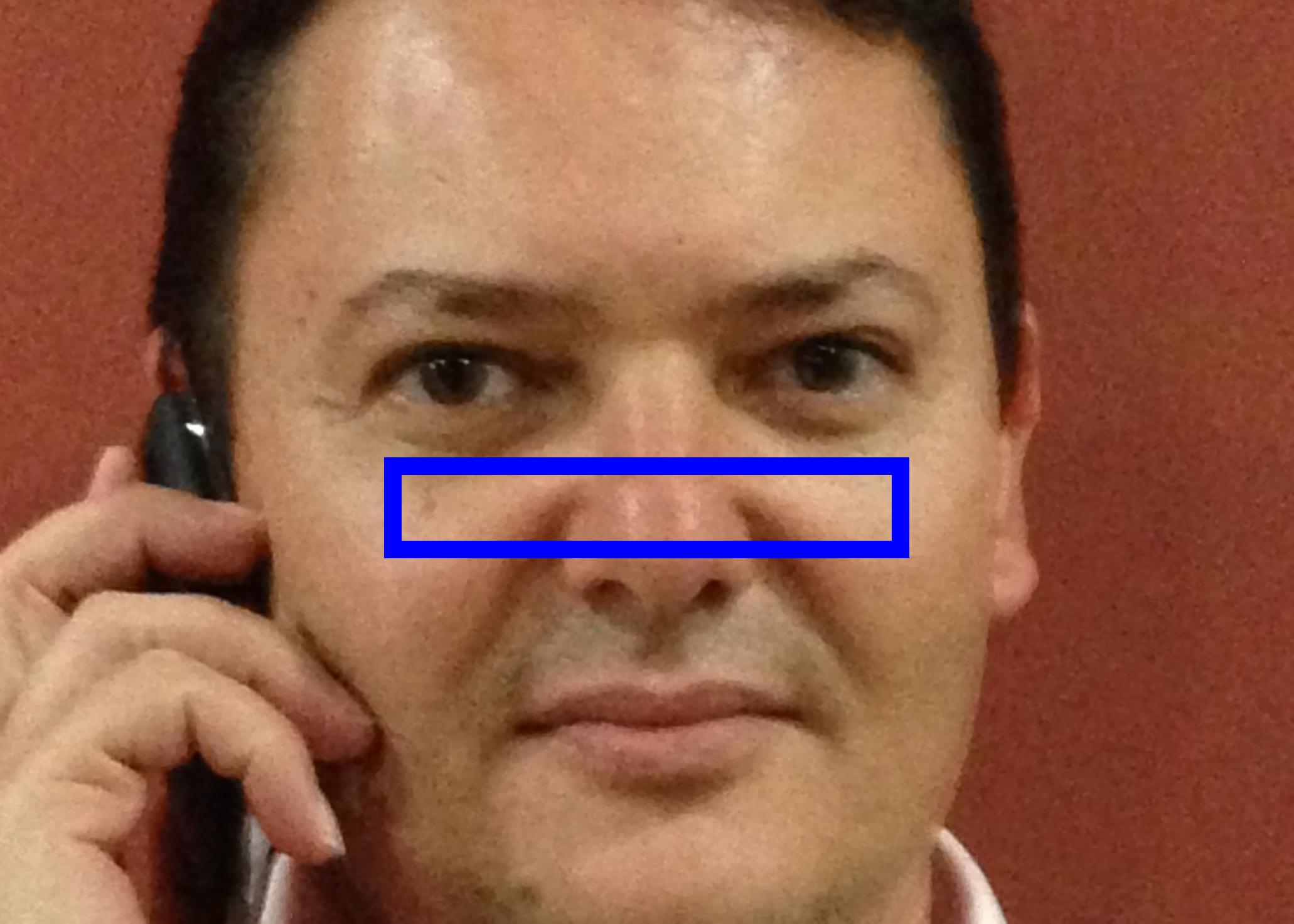} \label{fig:faceregcom}}
  \qquad
  \subfigure[Binary (with phone)]{\includegraphics[width=60px]{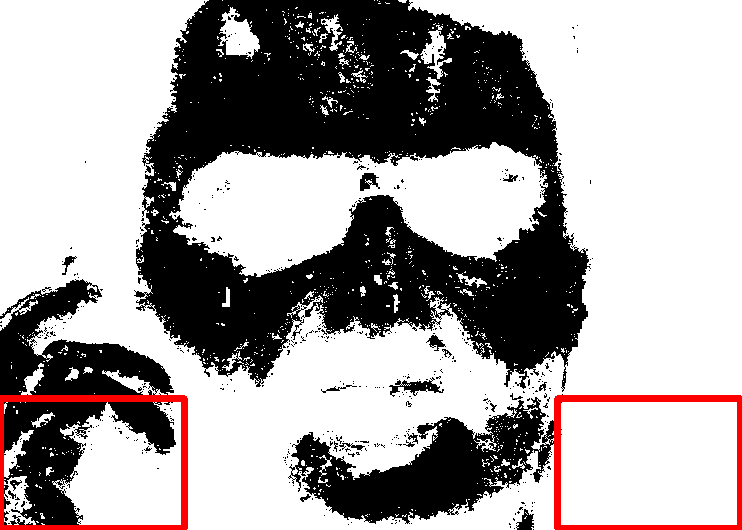} \label{fig:regcom}}
  \qquad
  \subfigure[Crop (no phone)]{\includegraphics[width=60px]{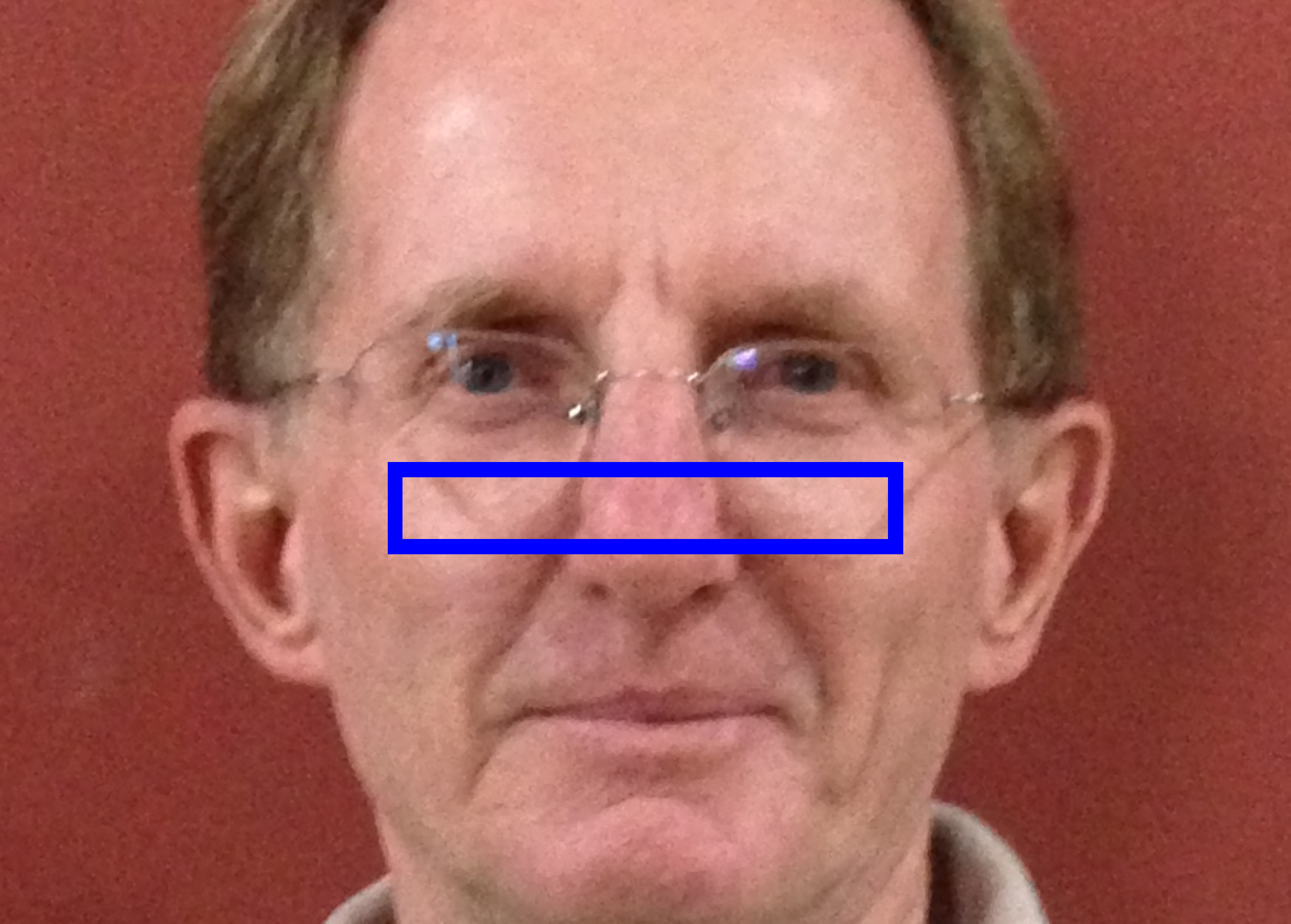} \label{fig:faceregsem}}
  \qquad
  \subfigure[Binary (no phone)]{\includegraphics[width=60px]{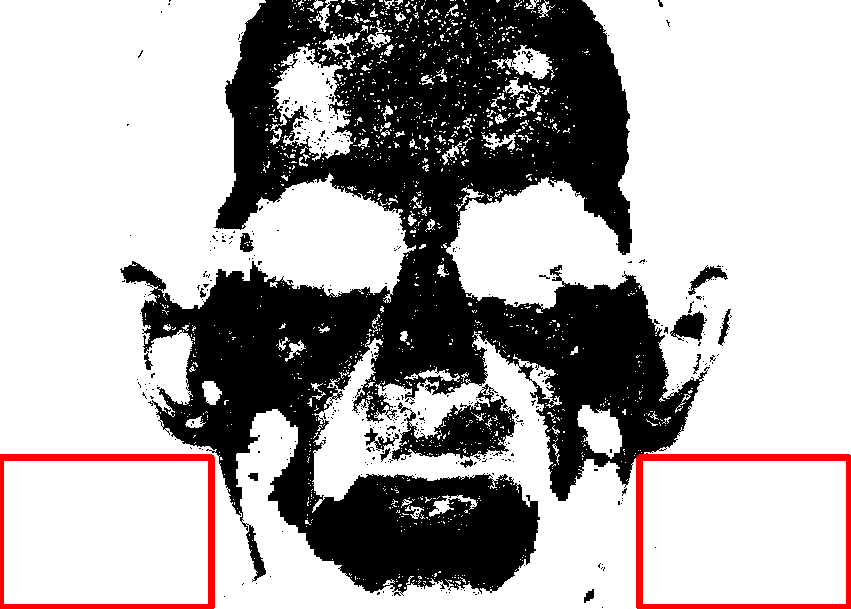} \label{fig:regsem}}
  \caption{The images (a) and (c) are examples of the driver face region (preprocessing), the rectangles (blue) are samples of skin for segmentation. Segmentation results are shown at images (b) and (d), the regions (red) where the pixels of the hand/arm are counted.}
  \label{fig:reg}
\end{figure}

$PH$ is obtained by counting the skin pixels in two regions of the image, as shown in red rectangles of Figures~\ref{fig:regcom} and \ref{fig:regsem}. These regions have the same size, 25\% of the height and 25\% of the width of the image and are in the bottom left and bottom right of the image. The attribute refers to the count of pixels in the Region 1 ($R_1$) and the pixels in Region 2 ($R_2$), dividing by the total of pixels ($T$). The formula is expressed in the Equation~\ref{eq:pm}.

\begin{equation}
 \label{eq:pm}
 PH = \frac{R_1 + R_2}{T}
\end{equation} 

$MI$, in turn, is calculated using the i\-ner\-tia moment of Hu (first moment of Hu \cite{hu1962}). It measures the pixels dispersion in the image. $MI$ results in a value nears to an object on different scales. Objects with different shapes have $MI$ values ​​far between.

General moment is calculated with Equation~\ref{eq:momentoRegular}, where $pq$ is called the order of the moment, $f(x,y)$ is the intensity of \textit{pixel} ($0$ or $1$ in binary images) at position $(x, y)$, $n_x$ and $n_y$ are the width and height of the image, respectively. The center of gravity or centroid of the image $(x_c, y_c)$ is defined by $(\frac{m_{10}}{m_{00}}, \frac{m_{01}}{m_{00}})$, and $m_{00}$ is area of the object for binary images. 
Central moments ($\mu_{pq}$) are defined by Equation~\ref{eq:momentoCentral}, where the centroid of the image is used in the calculation of the moment, and then gets invariance to location and orientation. Moments invariant to scale are obtained by normalizing as Equation~\ref{eq:momentoNormalizado}. $MI$ is defined, finally, by Equation~\ref{eq:momentoInercia}.

\begin{equation}
 \label{eq:momentoRegular}
 m_{pq} = \sum_{x = 1}^{n_x} \sum_{y = 1}^{n_y} x^p y^q f(x, y)
\end{equation}

\begin{equation}
 \label{eq:momentoCentral}
 \mu_{pq} = \sum_{x = 1}^{n_x} \sum_{y = 1}^{n_y} (x - x_c)^p (y - y_c)^q f(x, y)
\end{equation}

\begin{equation}
 \label{eq:momentoNormalizado}
 \eta_{pq} = \frac{\mu_{pq}}{\mu_{00}^{(1 + \frac{n + q}{2})}}
\end{equation}

\begin{equation}
 \label{eq:momentoInercia}
 MI = \eta_{20} + \eta_{02}
\end{equation}

\begin{figure}[ht!]
\center
  \subfigure[][MI=0.162]{\includegraphics[width=40px]{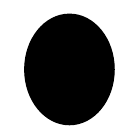}}
  \hspace{0px}
  \subfigure[][MI=0.162]{\includegraphics[width=40px]{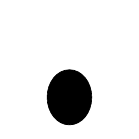}}
  \hspace{0px}
  \subfigure[][MI=0.170]{\includegraphics[width=40px]{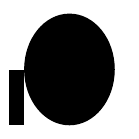}}
  \hspace{0px}
  \subfigure[][MI=0.170]{\includegraphics[width=40px]{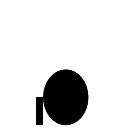}}
  \hspace{0px}
  \subfigure[][MI=0.170]{\includegraphics[width=40px]{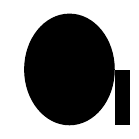}}
  \hspace{0px}
  \subfigure[][MI=0.170]{\includegraphics[width=40px]{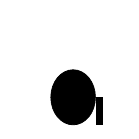}}
  \hspace{0px}
  \subfigure[][MI=0.166]{\includegraphics[width=40px]{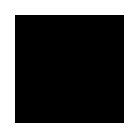}}
  \hspace{0px}
  \subfigure[][MI=0.166]{\includegraphics[width=40px]{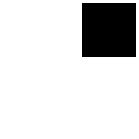}}
  
  \caption{Sample images and their Moments of Inertia (MI) calculated.}
  \label{fig:mi}
\end{figure}

Figure~\ref{fig:mi} shows some examples of the $MI$ calculations. The use of $MI$ aims to observe different standards for people with and without cell phone in the segmented image.

\section{\uppercase{Experiments}}
\label{sec:experimentos}

\begin{figure}[ht!]
\center
  \subfigure[][]{\includegraphics[width=40px]{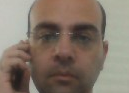}}
  \hspace{-3px}
  \subfigure[][]{\includegraphics[width=40px]{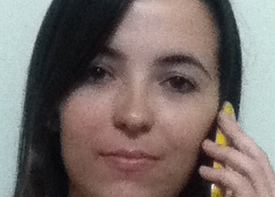}}
  \hspace{-3px}
  \subfigure[][]{\includegraphics[width=40px]{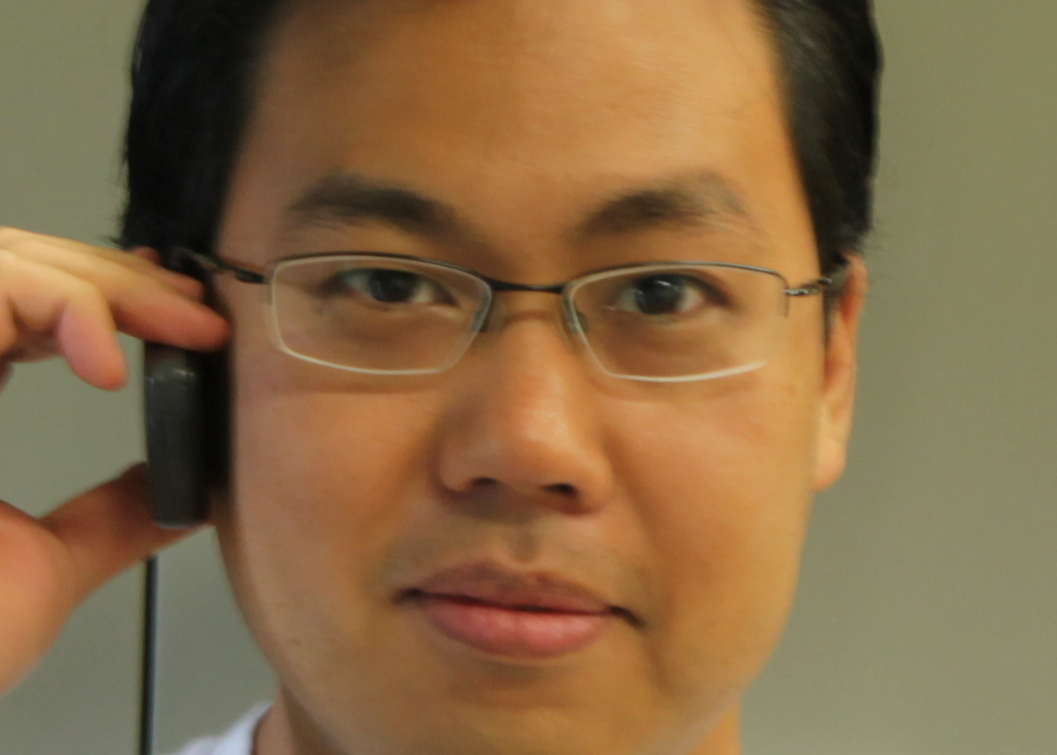}}
  \hspace{-3px}
  \subfigure[][]{\includegraphics[width=40px]{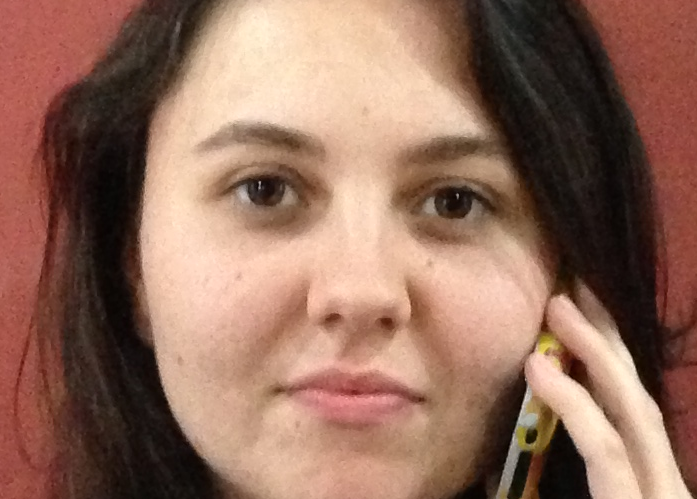}}
  \hspace{-3px}
  \subfigure[][]{\includegraphics[width=40px]{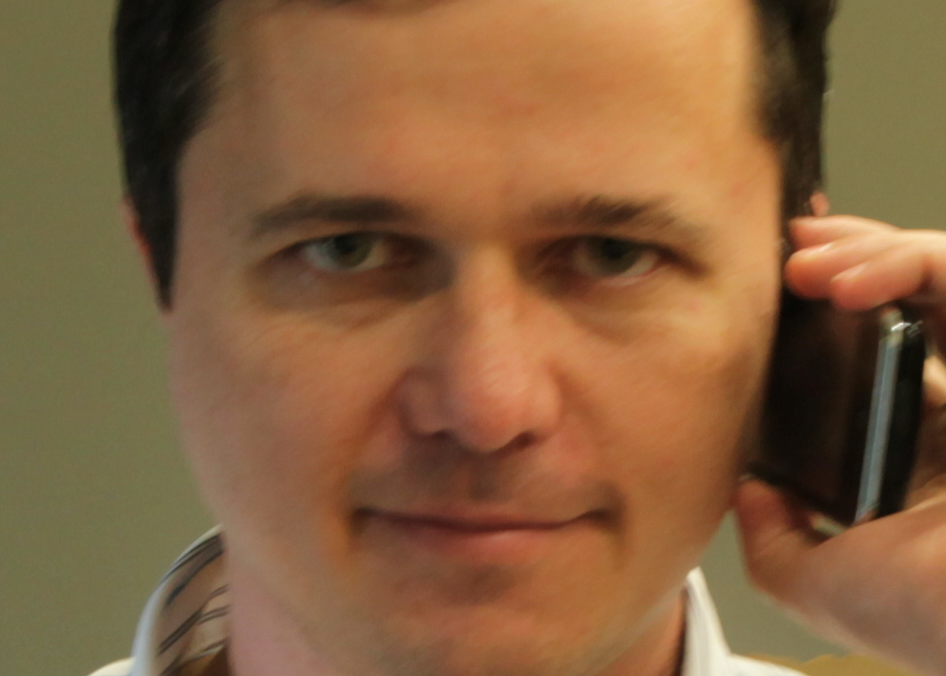}}

\setcounter{subfigure}{0} 
  \center
  \subfigure[][]{\includegraphics[width=40px]{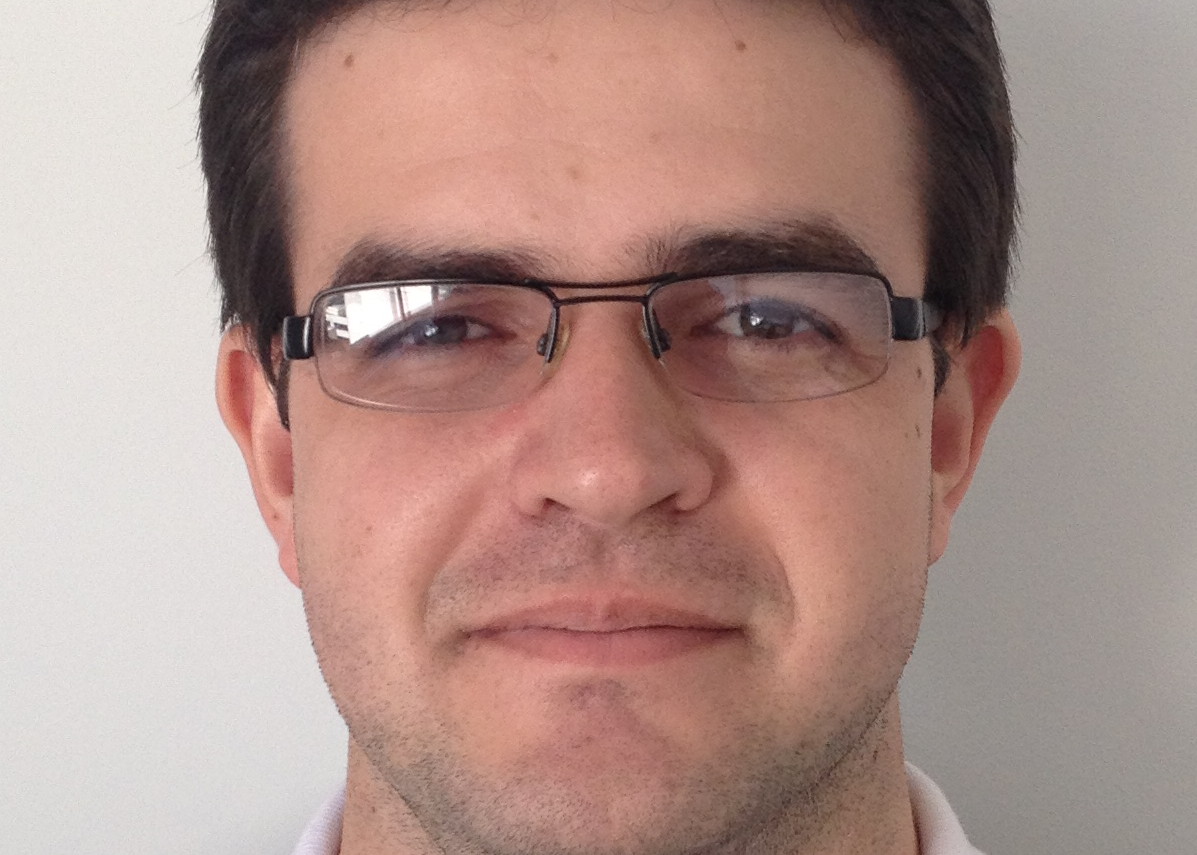}}
  \hspace{-3px}
  \subfigure[][]{\includegraphics[width=40px]{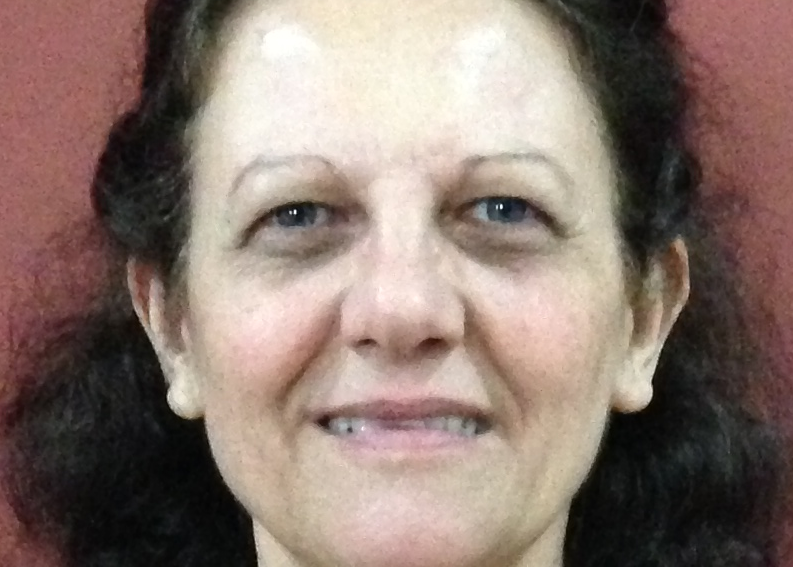}}
  \hspace{-3px}
  \subfigure[][]{\includegraphics[width=40px]{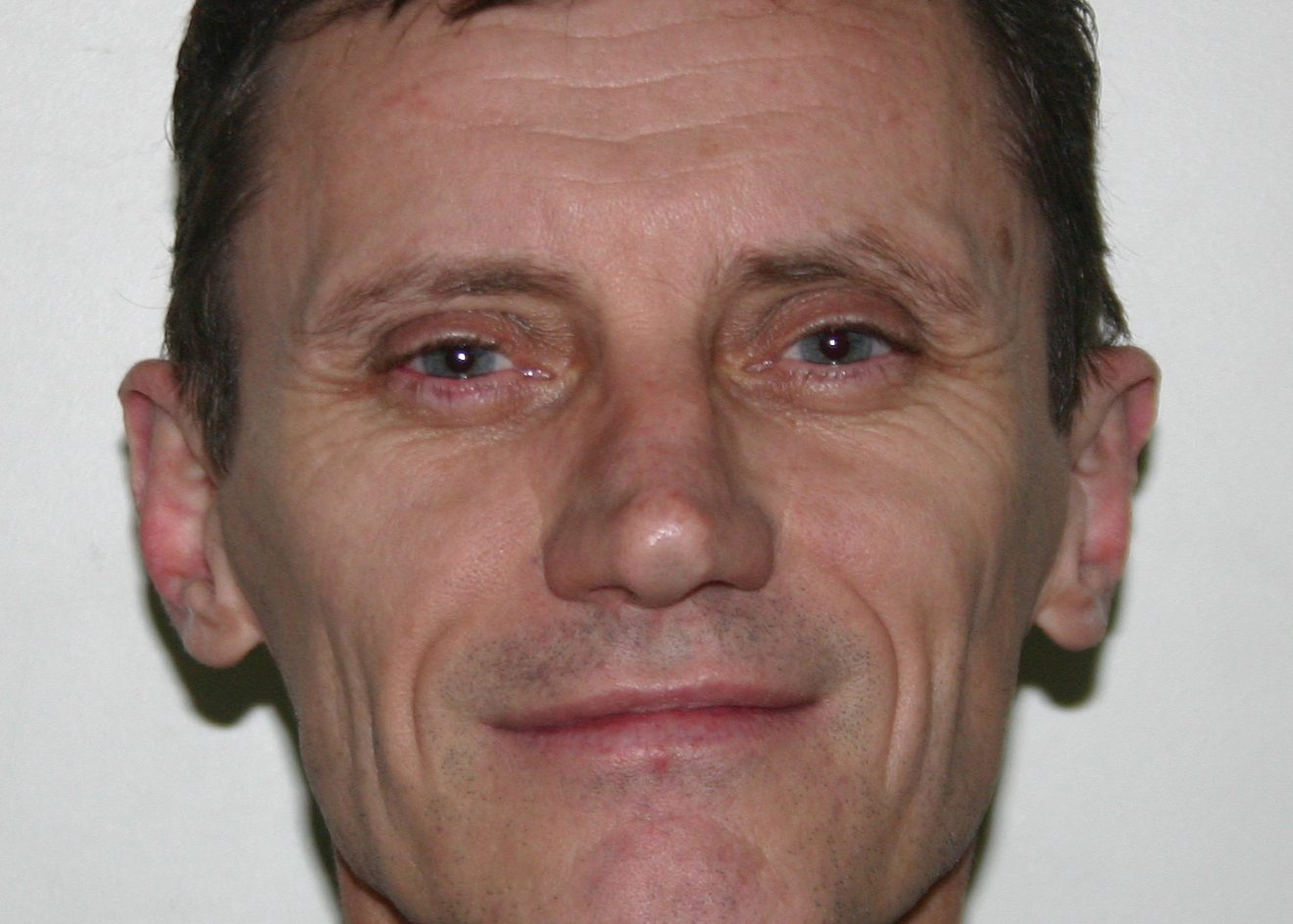}}
  \hspace{-3px}
  \subfigure[][]{\includegraphics[width=40px]{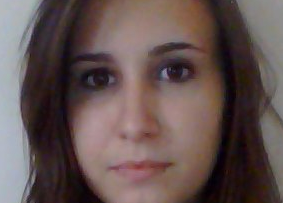}}
  \hspace{-3px}
  \subfigure[][]{\includegraphics[width=40px]{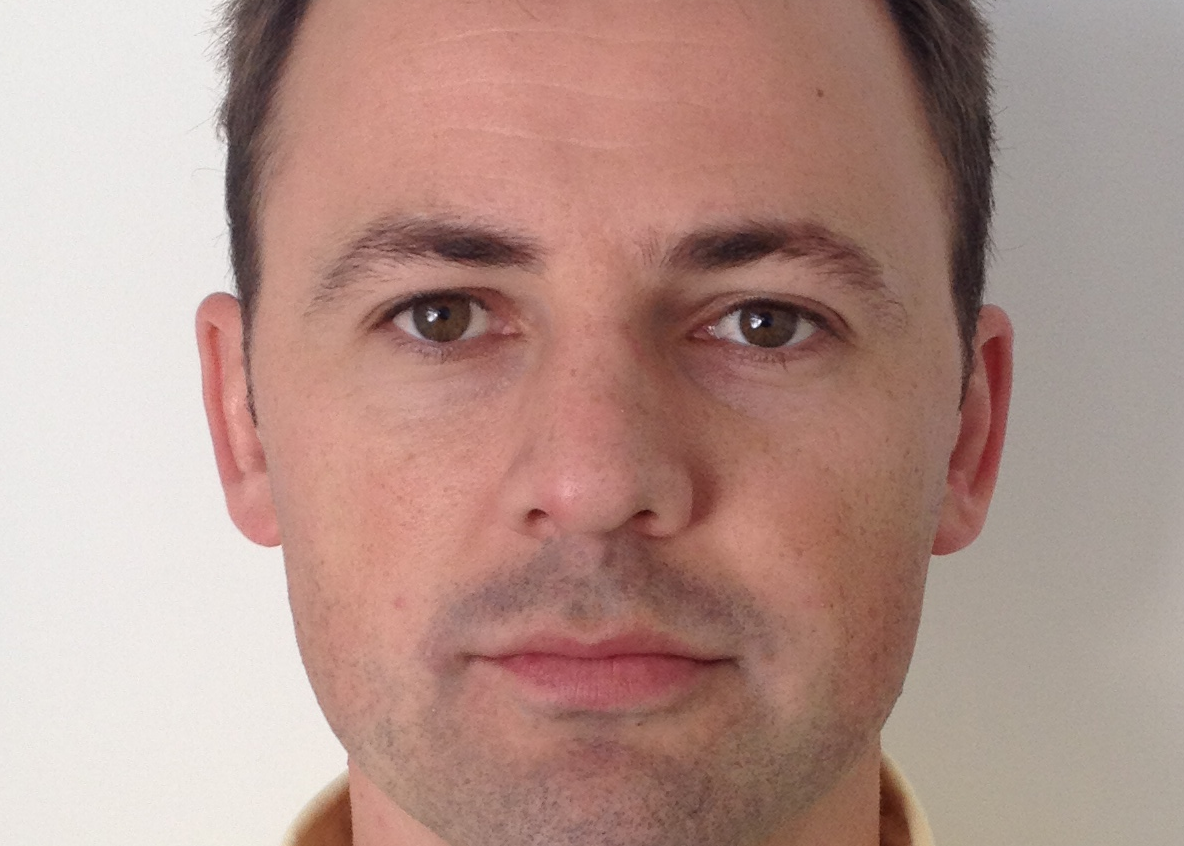}}
 
  \caption{Sample images for the set of images. Positive images are in first line. Negative images are in the second line.}
  \label{fig:banco}
\end{figure}

Experiments were performed on one set of i\-ma\-ges, with 100 positive images (people with phone) and the other 100 negative images (no phone). All images are frontal. In Figure~\ref{fig:banco}, sample images for the set of images are exemplified.

SVM is used as classification technique to the system. All tests have the same set of features and \textit{cross-validation} \cite{kohavi1995} is applied with 9 datasets from the initial set. Genetic Algorithm or GA\footnote{The library GALib version 2.4.7 is used (available in {\scriptsize \texttt{http://lancet.mit.edu/ga}}).} \cite{goldberg1989} is used to find the parameters ($nu$, $coef0$, $degree$, and $\gamma$ for SVM) for maximum accuracy of classification system.


GA is inspired by evolutionary biology, where the evolution of a population of individuals (candidate solutions) is simulated over several generations (iterations) in search of the best individual. GA parameters, empirically defined, used on experiments: $20$ individuals, $10,000$ generations, crossover $80\%$, mutation $5\%$, and tournament as selector. The initial po\-pu\-la\-tion is randomly (random parameters) and subsequent generations are the evolution result. The classifier is trained with parameters (genetic code) of each individual using binary encoding of 116 bits. Finally, the parameters of the individual that results in higher accuracy (fitness scores) for the classifier are adopted. GA was performed three times for each SVM kernel defined in Section~\ref{sec:svm}. The column ``Parameters'' of Tables~\ref{tab:svm} shows the best parameters found for SVM.

SVM is tested on kernels: Linear, Polynomial, RBF and Sigmoid (Section~\ref{sec:svm}). The \textit{kernel} that has a highest average accuracy is the Polynomial, reaching a rate of $91.57\%$ with images of training set. Table~\ref{tab:svm} shows the results of the tests, parameters and accuracy of each kernel.

\begin{table}[h!]
\caption{Accuracy of SVM kernels.}\label{tab:svm}
\begin{center}
\begin{tabular}{m{1.5cm}| >{\raggedright}p{2.5cm} | m{1cm} | m{0.7cm} }\hline
 & & \multicolumn{2}{ b{1.7cm} }{Accuracy (cross-validation)}\\\cline{3-4}
\multicolumn{1}{ b{1.4cm} | }{Kernel} & \multicolumn{1}{ b{2.5cm} | }{Parameters} & \multicolumn{1}{  b{1cm} | }{Average} & $\sigma$\\\hline
Linear & $nu=0.29$ & $91.06\%$ & $\pm6.90$\\\hline
Polynomial & $nu=0.30$ $coef0=4761.00$ $degree=0.25$ $\gamma=5795.48$ & $91.57\%$ & $\pm5.58$\\\hline
RBF & $nu=0.32$ $\gamma=0.36$ & $91.06\%$ & $\pm6.56$\\\hline
Sigmoid & $nu=0.23$ $coef0=1.82$ $\gamma=22.52$ & $89.06\%$ & $\pm8.79$\\\hline
\end{tabular}
\end{center}
\end{table}

The final tests are done with five videos\footnote{\label{videos}See the videos on the link:\\ {\scriptsize\texttt{http://www.youtube.com/channel/UCvwDU26FDAv1xO000AKrX0w}}} in real environments with the same driver. All videos have a variable frame rate. The average frame rate is $15$ FPS. The resolution is $320\times240$ pixels for all. The container format is 3GPP Media Release 4 for them. The specific information about the videos are in Table~\ref{tab:videos}.

\begin{table}[h!]
\caption{Information about the videos.}\label{tab:videos}
\begin{center}
\begin{tabular}{p{0.3cm}|p{1.6cm}|p{1.1cm}| >{\raggedleft}p{1.2cm} | >{\raggedleft}p{0.9cm}} \hline
\# & Weather & Time & Duration & Frames \tabularnewline\hline
V1 & Overcast & Noon & $735$ s & $11,005$ \tabularnewline\hline
V2 & Mainly sun & Noon & $1,779$ s & $26,686$ \tabularnewline\hline
V3 & Sunny & Noon & $1,437$ s & $21,543$ \tabularnewline\hline
V4 & Sunny & Dawn & $570$ s & $8,526$ \tabularnewline\hline
V5 & Sunny & Late afternoon & $630$ s & $9,431$ \tabularnewline\hline
\end{tabular}
\end{center}
\end{table}

\begin{figure}[ht!]
\center
  \subfigure[][]{\includegraphics[width=69px]{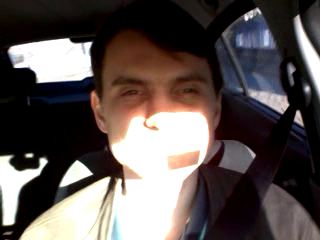}}
  \hspace{-3px}
  \subfigure[][]{\includegraphics[width=69px]{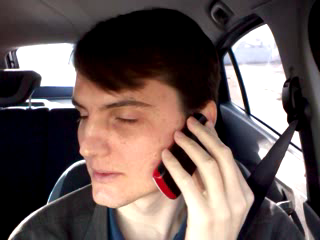}}
  \hspace{-3px}
  \subfigure[][]{\includegraphics[width=69px]{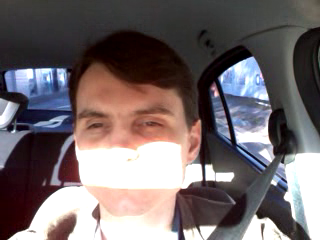}}
\center
  \subfigure[][]{\includegraphics[width=69px]{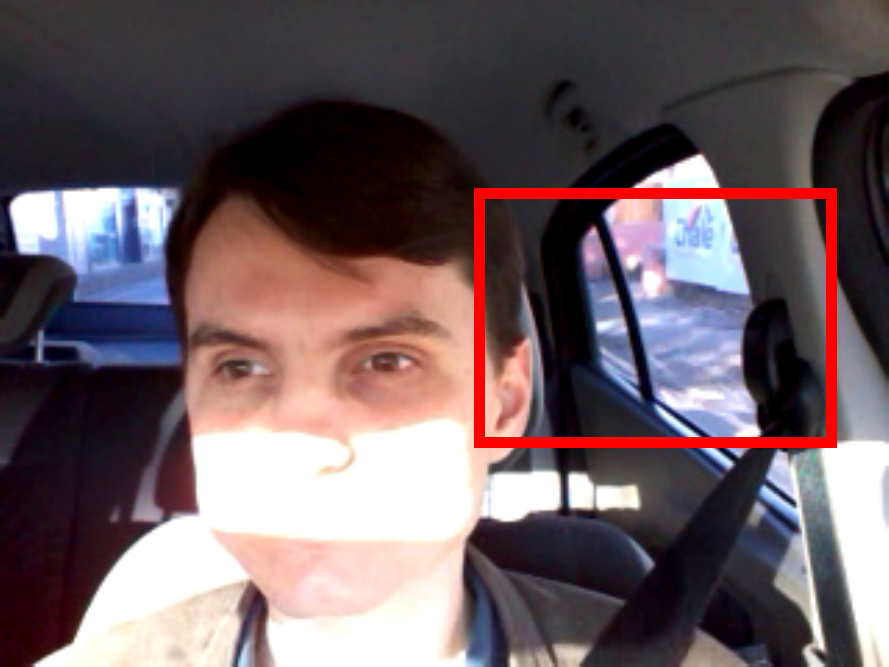}}
  \hspace{-3px}
  \subfigure[][]{\includegraphics[width=69px]{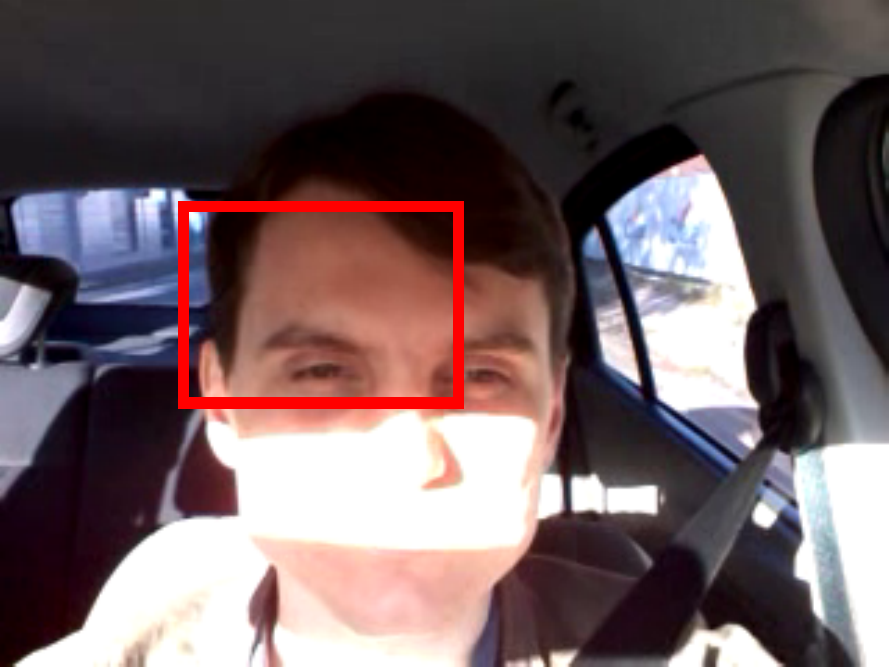}}
  \hspace{-3px}
  \subfigure[][]{\includegraphics[width=69px]{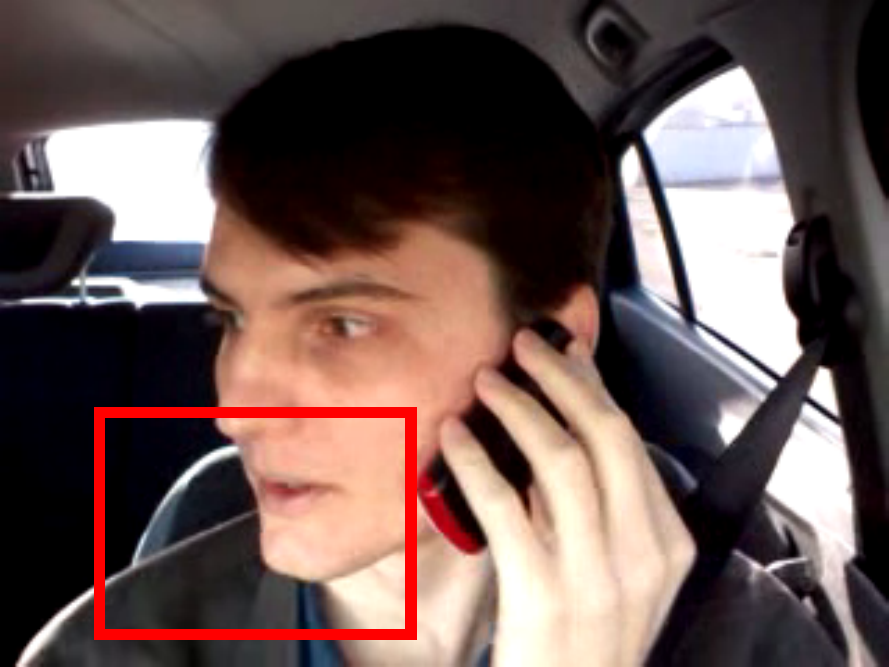}}
 
  \caption{Sample frames for preprocessing problem. The frames in (a), (b), and (c) are examples for driver not found. The frames in (d), (e), and (f) are examples for driver wrong, and the red rectangle shows where the driver was found for these images.}
  \label{fig:preprocessprob}
\end{figure}

Some mistakes were observed in the frames in preprocessing step (Section~\ref{sec:prepro}). The Table~\ref{tab:preprocesso} shows the problems encountered. In frames reported as ``not found'' the driver cannot be found. The frames of the column ``wrong'' are false positives, ie, the region declared as a driver is wrong. The preprocessing algorithm was worst for Video 4 with rate error $24.20\%$. Some frames with preprocessing problem are shown in Figure~\ref{fig:preprocessprob}. The best video for this step was Video 3 with rate error $3.31\%$. The rate error average for all videos' frames is $7.42\%$. The frames with pre\-pro\-ces\-sing errors were excluded from following experiments.

\begin{table}[h!]
\caption{Error while searching for Driver (preprocessing).}\label{tab:preprocesso}
\begin{center}
\begin{tabular}{l|r|r|r|m{1.5cm}}\hline
\# & Not found & Wrong & Total & Frame rate error\\\hline
V1 & $237$ & $273$ & $510$ & $4.63\%$\\\hline
V2 & $675$ & $974$ & $1,649$ & $6.18\%$\\\hline
V3 & $356$ & $358$ & $714$ & $3.31\%$\\\hline
V4 & $1,280$ & $783$ & $2,063$ & $24.20\%$\\\hline
V5 & $533$ & $259$ & $792$ & $8.40\%$\\\hline
\end{tabular}
\end{center}
\end{table}

\begin{figure}[h!]
  \centering
  \includegraphics[width=220px]{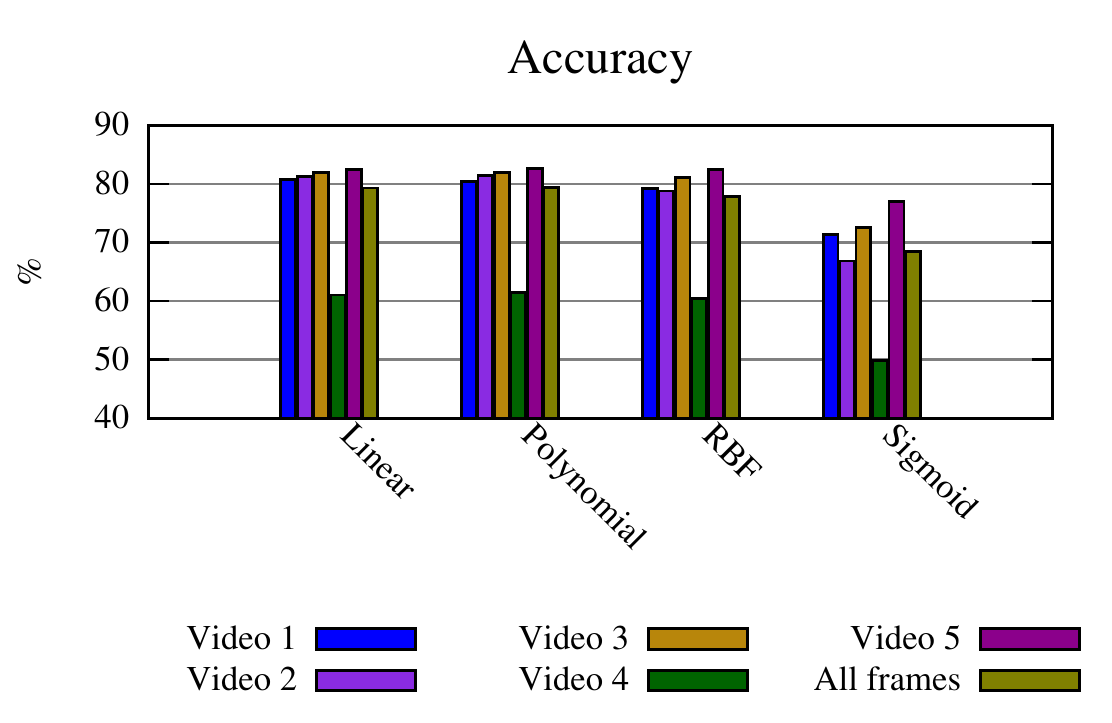}
  \caption{Accuracy for the kernels by frame.}
  \label{fig:grafkernelsacuracia}
\end{figure}

Each frame of the video was segmented (Section~\ref{sec:segment}), extracted features (Section~\ref{sec:carac}), and classified by kernels. 
Figure~\ref{fig:grafkernelsacuracia} shows the results for each combination of video/kernel and the last line shows the average accuracy values for frames of all videos.
The polynomial kernel was the best on the tests, it obtained an accuracy of 79.36\%. 
For the next experiment we opted for the polynomial kernel.

We performed time analysis splitting the videos in pe\-riods of 3 seconds. The Table~\ref{tab:videoperiodoacuracia} shows the periods quantity by video.

\begin{table}[h!]
\caption{Number of periods by video.}\label{tab:videoperiodoacuracia}
\begin{center}
\begin{tabular}{l|l|l|l|l|l}\hline
Videos & V1 & V2 & V3 & V4 & V5\\\hline
Periods & $245$ & $594$ & $479$ & $190$ & $210$\\\hline
\end{tabular}
\end{center}
\end{table}

\begin{figure}[h!]
  \centering
  \center
  \subfigure[][V1]{\includegraphics[width=160px]{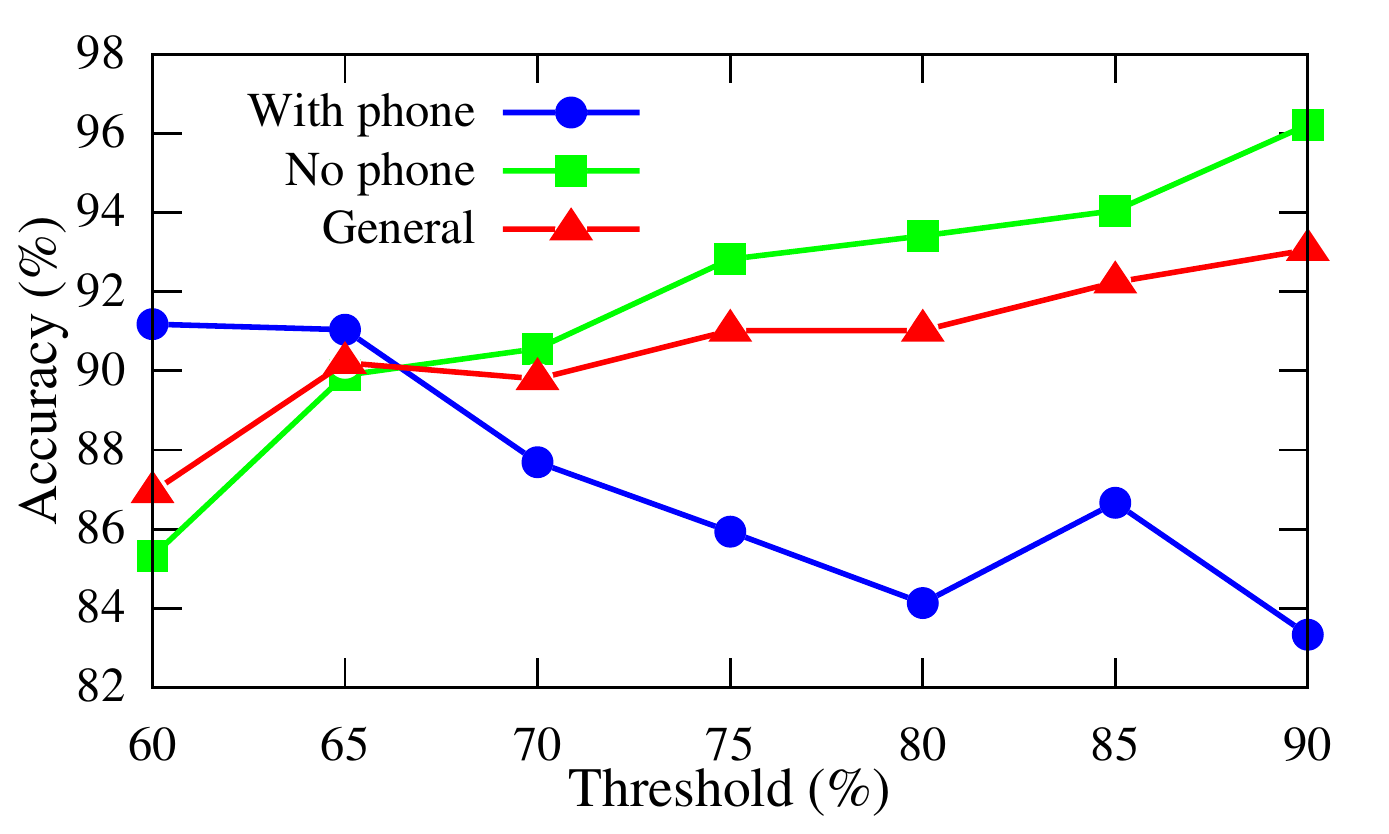}\label{fig:grafcortevideov1}}
  \subfigure[][V2]{\includegraphics[width=160px]{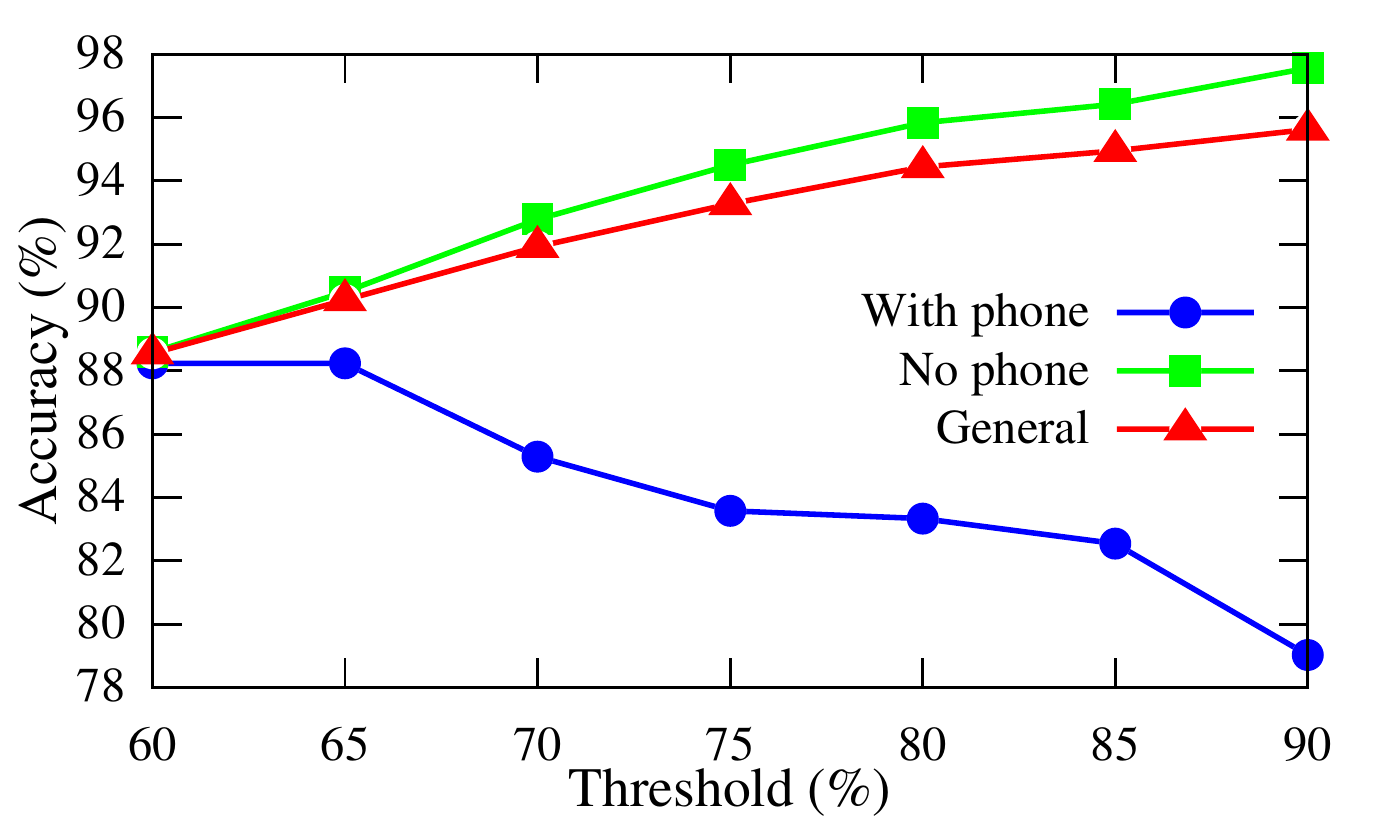}\label{fig:grafcortevideov2}}
  \subfigure[][V3]{\includegraphics[width=160px]{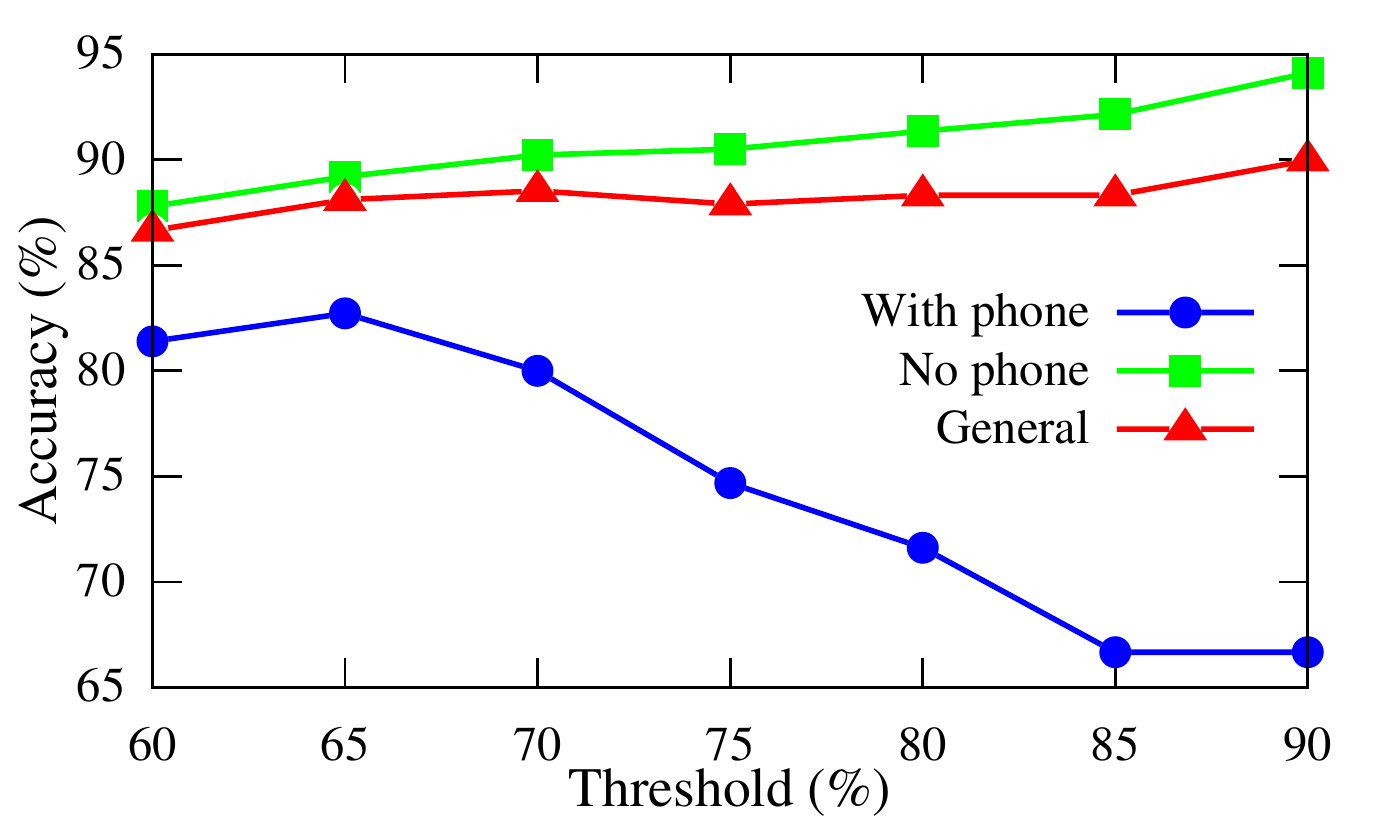}\label{fig:grafcortevideov3}}
  \subfigure[][V4]{\includegraphics[width=160px]{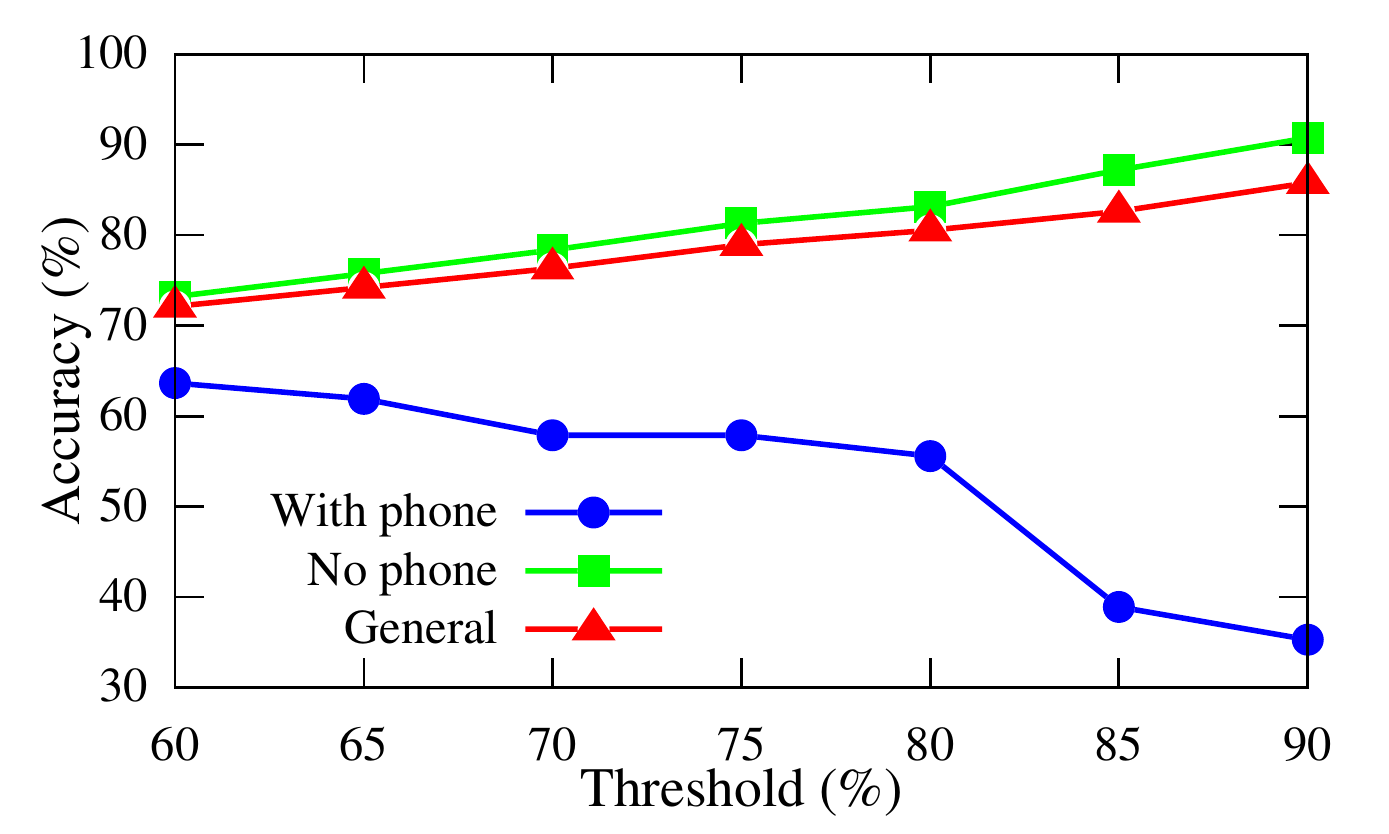}\label{fig:grafcortevideov4}}
  \subfigure[][V5]{\includegraphics[width=160px]{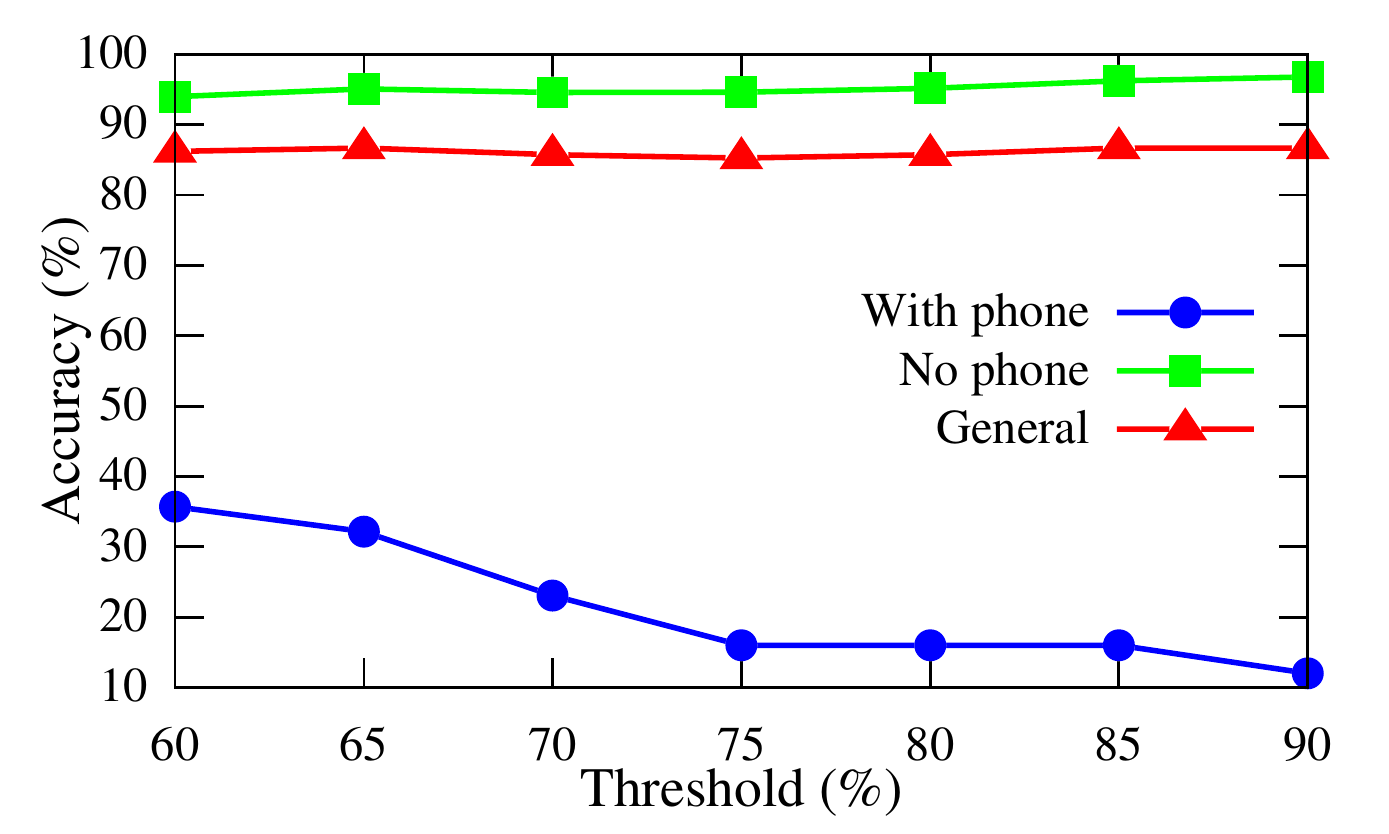}\label{fig:grafcortevideov5}}
  \caption{Accuracy for polynomial kernel by period and video.}
  \label{fig:grafcortevideo}
\end{figure}

Detection of cell phone usage happens when the period has the frames rate classified individually with cell phone more or equal of a \textit{threshold}.
The \textit{threshold} values $​​60\%$, $65\%$, $70\%$, $75\%$, $80\%$, 85\%, and 90\% were tested with the videos. The accuracy graphs are shown in Figure~\ref{fig:grafcortevideo}. The columns ``With phone'', ``No phone'', and ``General'' represent the accuracy obtained for frame with cell phone, no cell phone, and in general, respectively.

\begin{figure}[ht!]
\center
  \subfigure[][]{\includegraphics[width=69px]{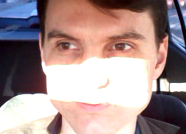}}
  \hspace{-3px}
  \subfigure[][]{\includegraphics[width=69px]{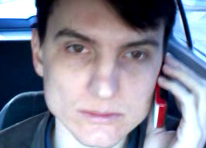}}
  \hspace{-3px}
  \subfigure[][]{\includegraphics[width=69px]{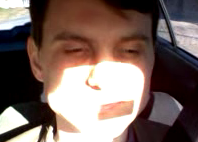}}

\setcounter{subfigure}{0}   
\center
  \subfigure[][]{\includegraphics[width=69px]{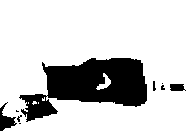}}
  \hspace{-3px}
  \subfigure[][]{\includegraphics[width=69px]{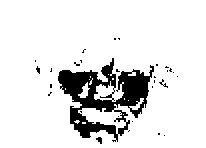}}
  \hspace{-3px}
  \subfigure[][]{\includegraphics[width=69px]{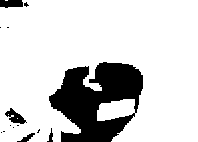}}
 
  \caption{Sample frames for segmentation problem. In first line are examples for driver's face. In second line are examples of segmentation.}
  \label{fig:segmentaprob}
\end{figure}

Figure~\ref{fig:grafcortevideov4} shows a larger distance between accuracy ``with phone'' and ``no phone'' for the Videos 4 and 5. This difference is caused by problems with preprocessing (Figure~\ref{fig:preprocessprob}), and segmentation. The preprocessing problems causes to decrease number of frames analyzed in some periods, thus the classification is impaired. Segmentation's problems are caused by the incidence of sunlight (close to twilight mainly) on some regions of the driver's face and the inner parts of the vehicle. The sunlight changes the components of pixels. Incorrect segmentation compromises the feature extraction can lead to misclassification of frames. Figure~\ref{fig:segmentaprob} exemplifies the segmentation pro\-blems.

\begin{figure}[h!]
  \centering
  \includegraphics[width=170px]{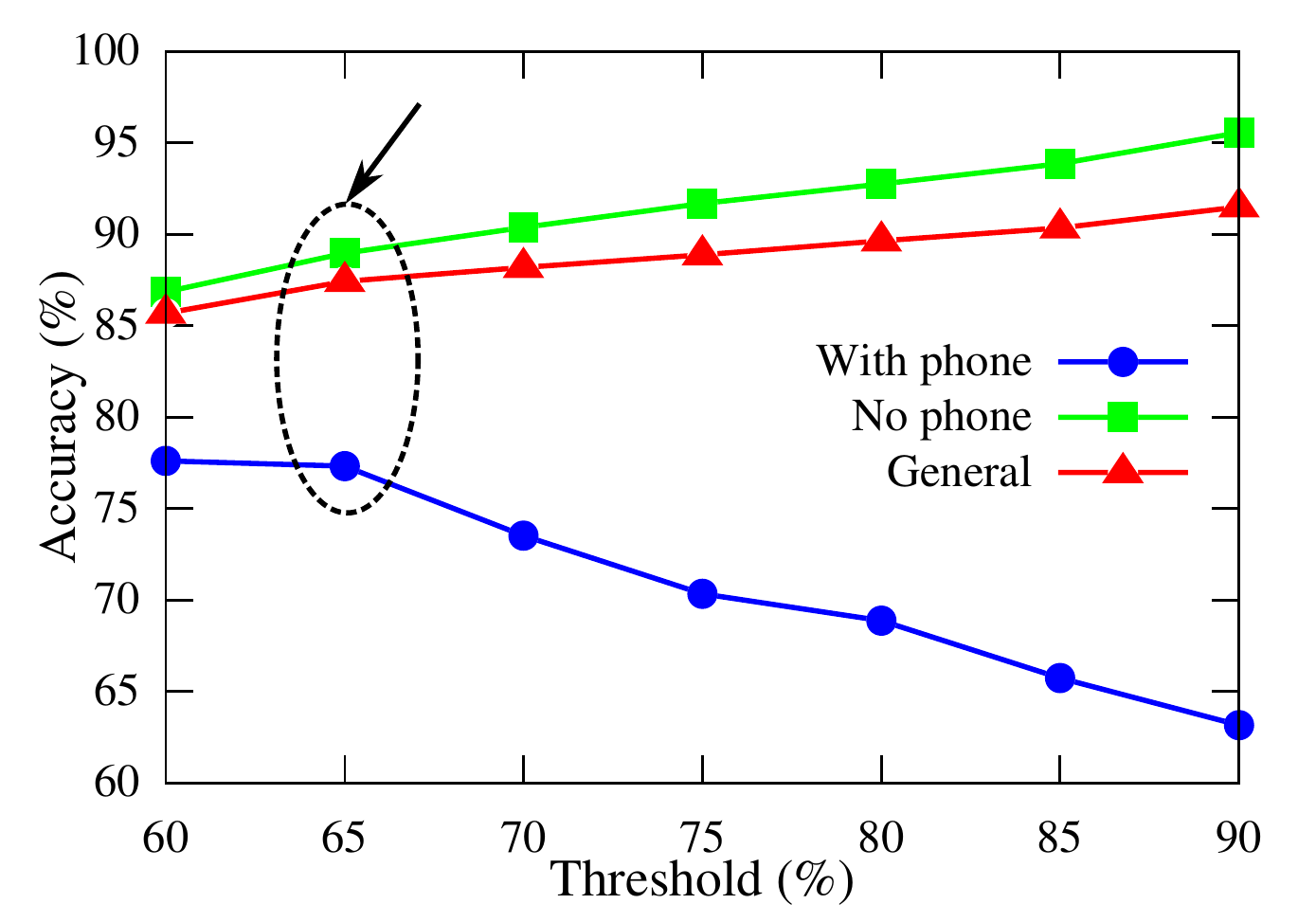}
  \caption{Graph of accuracy average (polynomial kernel) by period for all videos of threshold by phone, No phone, and General. The ellipse shows the best threshold.}
  \label{fig:grafcorte}
\end{figure}

Figure~\ref{fig:grafcorte} presents the mean accuracy: ``with phone'', ``no phone'', and in general, for videos at each threshold. 
The accuracy of detecting cell phone is greatest with thresholds $60\%$ and $65\%$. The accuracy ``no phone'' for threshold $65\%$ is better than $60\%$. Thus, the threshold of $65\%$ is more advantageous for videos. This threshold results in accuracy for ``with phone'' $77.33\%$, ``no phone'' $88.97\%$, and in general of $87.43\%$ at three videos.

\subsection{Real time system}\label{sec:sisTempoReal}


The five videos\footnote{\label{videos}See the videos with real time classification on the link {\scriptsize\texttt{http://www.youtube.com/channel/UCvwDU26FDAv1xO000AKrX0w}}} were simulated in real time on a computer Intel i5 2.3GHz CPU, 6.0GB RAM and operating system Lubuntu 12.04. SVM/Polynomial classifier was used, and training with the images set. The videos were used with a resolution of $320\times240$ pi\-xels.


The system uses parallel processing (\textit{multi-thread}), as follow one thread for image acquisition, the frames are processed for four threads, and one thread to display the result. Thus four frames can be processed simultaneously. It shows able to process up to $6.4$ frames per second (FPS), however, to avoid bottlenecks is adopted the rate $6$ FPS for its execution.


The input image is changed to illustrate the result of processing in the output of system. At left center is added a vertical progress bar or historical per\-cen\-ta\-ge for frames ``with phone'' in the previous period (3 second). When the frames percentage is between $0\%$ and $40\%$ green color is used to fill, the yellow color is used between $40\%$ and $65\%$, and red color above or equal to $65\%$. Red indicates a risk situation. For real situation should be started a beep. Figure 1 shows a sequence of frames the system's output.


\begin{figure}[ht!]
  
  
  \center\vspace{-5px}
  \subfigure{\includegraphics[width=41px]{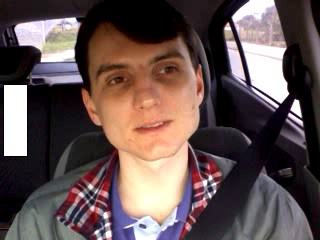}}
  \hspace{-4px}
  \subfigure{\includegraphics[width=41px]{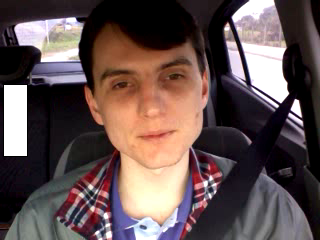}}
  \hspace{-4px}
  \subfigure{\includegraphics[width=41px]{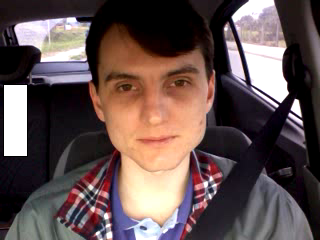}}
  \hspace{-4px}
  \subfigure{\includegraphics[width=41px]{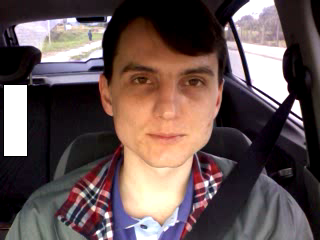}}
  \hspace{-4px}
  \subfigure{\includegraphics[width=41px]{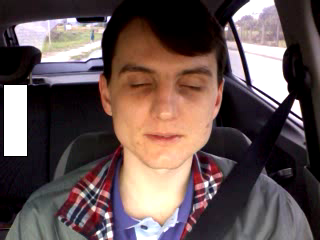}}
  
  \center\vspace{-5px}
  \subfigure{\includegraphics[width=41px]{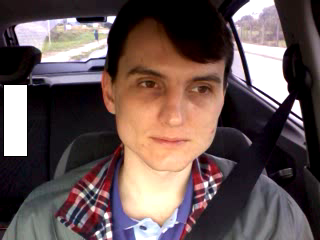}}
  \hspace{-4px}
  \subfigure{\includegraphics[width=41px]{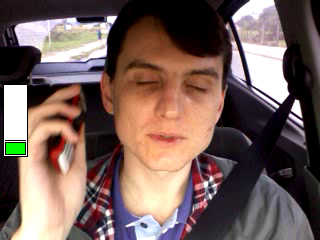}}
  \hspace{-4px}
  \subfigure{\includegraphics[width=41px]{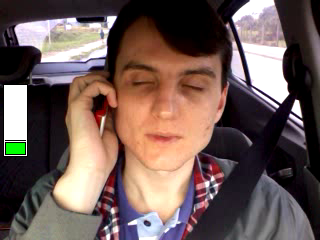}}
  \hspace{-4px}
  \subfigure{\includegraphics[width=41px]{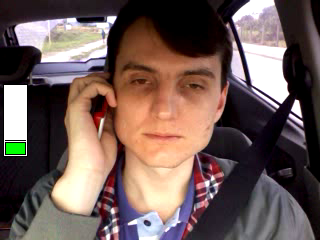}}
  \hspace{-4px}
  \subfigure{\includegraphics[width=41px]{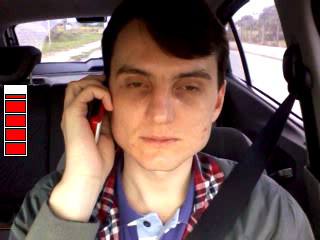}}
  
  \center\vspace{-5px}
  \subfigure{\includegraphics[width=41px]{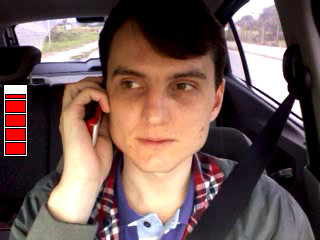}}
  \hspace{-4px}
  \subfigure{\includegraphics[width=41px]{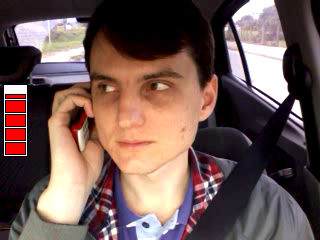}}
  \hspace{-4px}
  \subfigure{\includegraphics[width=41px]{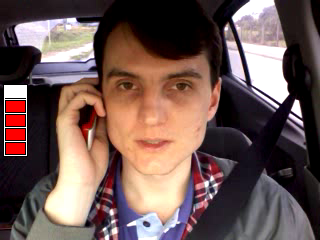}}
  \hspace{-4px}
  \subfigure{\includegraphics[width=41px]{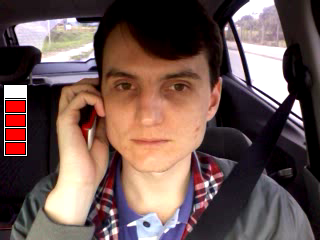}}
  \hspace{-4px}
  \subfigure{\includegraphics[width=41px]{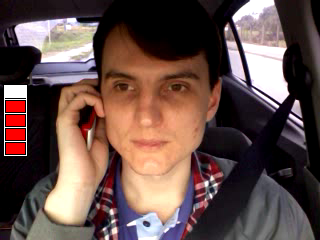}}
  
  
  \caption{The system's output sequence sample in real time with one frame for each second ($15$ seconds).}
  \label{fig:seqSaidaSistema}
\end{figure}

\section{\uppercase{Conclusion}}
\label{sec:conclusao}

This paper presented an algorithm which allows extraction features of an image to detect the use of cell phones by drivers in a car. Very little literature on this subject was observed (usually the focus is on drowsiness detection or analysis of external objects or pedestrians). The Table~\ref{tab:comparacao} compare related works with this work.

\begin{table}[h!]
\caption{Table comparing related works and this work}\label{tab:comparacao}
\begin{center}
\includegraphics[]{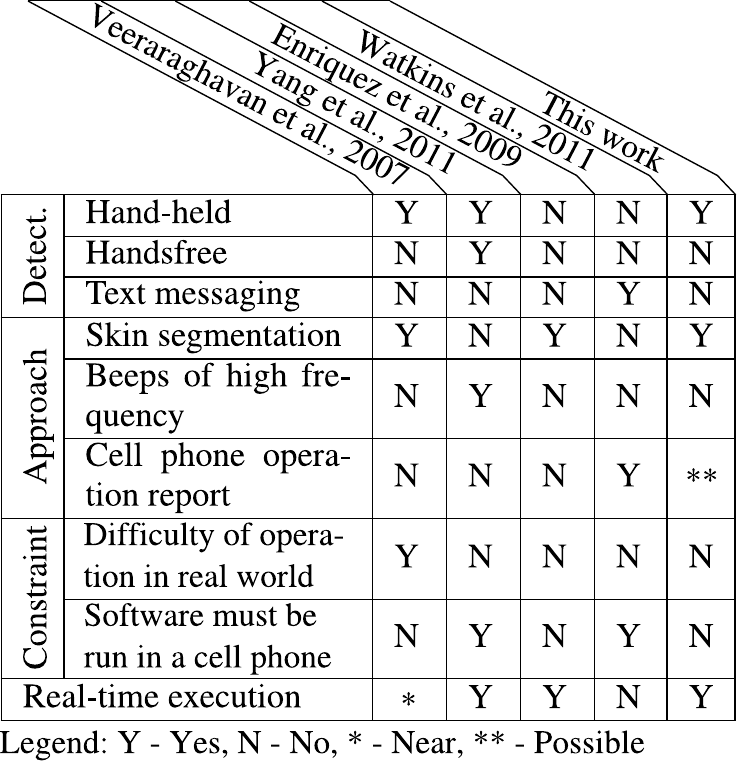}
\end{center}
\end{table}


SVM and its kernels were tested as candidates to solve the problem. The polynomial kernel (SVM) is the most advantageous classification system with a\-ve\-ra\-ge accuracy of $91.57\%$ for set of images analyzed. But, all the kernels tested have average accuracy statistically the same. GA found statistically similar parameters for all kernels.

Tests done on videos show that it is possible to use the image datasets for training classifiers in real si\-tua\-tions. Periods of 3 seconds were correctly classified at $87.43\%$ of cases. The segmentation algorithm tends to fail when the sunlight falls (Videos~4 and 5) at driver face and parts inside vehicle. This changes the components value for pixels of driver skin. Thus, the pixels of skin for face and hand are different. 

Enhanced cell phone use detection system is able to find ways to works better when the sunlight falls at the driver skin.
Another improvement is to check if the vehicle is moving and then execute the detection.
An intelligent warning signal can be created, i.e., more sensitive detection according to the speed increases. One way is to join the OpenXC Platform\footnote{It allows you to develop applications integrated with the vehicle (available in {\scriptsize \texttt{http://openxcplatform.com}})} on solution to get the real speed among other data. 

\section*{ACKNOWLEDGEMENTS}

We thank CAPES/DS for the financial support to Rafael Alceste Berri of Graduate Program in Applied Computing of Santa Catarina State University (UDESC), and PROBIC/UDESC to Elaine Girardi of Undergraduate program in Electrical Engineering.

\vfill
\bibliographystyle{apalike}

{\small
\begin{thebibliography}{}

\bibitem[Balbinot et~al., 2011]{balbinot2011}
Balbinot, A.~B., Zaro, M.~A., and Timm, M.~I. (2011).
\newblock Fun{\c{c}}{\~o}es psicol{\'o}gicas e cognitivas presentes no ato de
  dirigir e sua import{\^a}ncia para os motoristas no tr{\^a}nsito.
\newblock {\em Ci{\^e}ncias e Cogni{\c{c}}{\~a}o/Science and Cognition}, 16(2).

\bibitem[Dadgostar and Sarrafzadeh, 2006]{dadgostar2006}
Dadgostar, F. and Sarrafzadeh, A. (2006).
\newblock An adaptive real-time skin detector based on hue thresholding: A
  comparison on two motion tracking methods.
\newblock {\em Pattern Recognition Letters}, 27(12):1342--1352.

\bibitem[Drews et~al., 2008]{drews2008}
Drews, F.~A., Pasupathi, M., and Strayer, D.~L. (2008).
\newblock Passenger and cell phone conversations in simulated driving.
\newblock {\em Journal of Experimental Psychology: Applied}, 14(4):392.

\bibitem[Enriquez et~al., 2009]{enriquez2009}
Enriquez, I. J.~G., Bonilla, M. N.~I., and Cortes, J. M.~R. (2009).
\newblock Segmentacion de rostro por color de la piel aplicado a deteccion de
  somnolencia en el conductor.
\newblock {\em Congreso Nacional de Ingenieria Electronica del Golfo
  CONAGOLFO}, pages 67--72.

\bibitem[Goldberg, 1989]{goldberg1989}
Goldberg, D.~E. (1989).
\newblock {\em {Genetic Algorithms in Search, Optimization, and Machine
  Learning}}.
\newblock Addison-Wesley Professional, 1 edition.

\bibitem[Goodman et~al., 1997]{goodman1997}
Goodman, M., Benel, D., Lerner, N., Wierwille, W., Tijerina, L., and Bents, F.
  (1997).
\newblock {\em An investigation of the safety implications of wireless
  communications in vehicles}.
\newblock US Dept. of Transportation, National Highway Transportation Safety
  Administration.

\bibitem[Hearst et~al., 1998]{hearst1998}
Hearst, M.~A., Dumais, S.~T., Osman, E., Platt, J., and Sch{\"o}lkopf, B.
  (1998).
\newblock Support vector machines.
\newblock {\em Intelligent Systems and their Applications, IEEE}, 13(4):18--28.

\bibitem[Hu, 1962]{hu1962}
Hu, M.~K. (1962).
\newblock Visual pattern recognition by moment in\-va\-riants.
\newblock {\em Information Theory, IRE Transactions on}, 8(2):179--187.

\bibitem[Kohavi, 1995]{kohavi1995}
Kohavi, R. (1995).
\newblock A study of cross-validation and bootstrap for accuracy estimation and
  model selection.
\newblock In {\em International joint Conference on artificial intelligence},
  volume~14, pages 1137--1145. Lawrence Erlbaum Associates Ltd.

\bibitem[NHTSA, 2011]{nhtsa2011}
NHTSA (2011).
\newblock Driver electronic device use in 2010.
\newblock {\em Traffic Safety Facts - December 2011}, pages 1--8.

\bibitem[Peissner et~al., 2011]{peissner2011}
Peissner, M., Doebler, V., and Metze, F. (2011).
\newblock Can voice in\-te\-rac\-tion help reducing the level of distraction
  and prevent accidents?

\bibitem[Regan et~al., 2008]{regan2008}
Regan, M.~A., Lee, J.~D., and Young, K.~L. (2008).
\newblock {\em Driver distraction: Theory, effects, and mitigation}.
\newblock CRC.

\bibitem[Smith and Chang, 1996]{smith1996}
Smith, J.~R. and Chang, S.~F. (1996).
\newblock Tools and techniques for color image retrieval.
\newblock In {\em SPIE proceedings}, volume 2670, pages 1630--1639.

\bibitem[Stanimirova et~al., 2010]{stanimirova2010}
Stanimirova, I., {\"U}st{\"u}n, B., Cajka, T., Riddelova, K., Hajslova, J.,
  Buydens, L., and Walczak, B. (2010).
\newblock Tracing the geographical origin of honeys based on volatile compounds
  profiles assessment using pattern recognition techniques.
\newblock {\em Food Chemistry}, 118(1):171--176.

\bibitem[Strayer et~al., 2013]{strayer2013}
Strayer, D.~L., Cooper, J.~M., Turrill, J., Coleman, J., Medeiros-Ward, N., and
  Biondi, F. (2013).
\newblock Measuring cognitive distraction in the automobile.
\newblock {\em AAA Foundation for Traffic Safety - June 2013}, pages 1--34.

\bibitem[Strayer et~al., 2011]{strayer2011}
Strayer, D.~L., Watson, J.~M., and Drews, F.~A. (2011).
\newblock 2 cognitive distraction while multitasking in the automobile.
\newblock {\em Psychology of Learning and Motivation-Advances in Research and
  Theory}, 54:29.

\bibitem[Vapnik, 1995]{vapnik1995}
Vapnik, V. (1995).
\newblock {\em The nature of statistical learning theory}.
\newblock Springer-Verlag, New York.

\bibitem[Veeraraghavan et~al., 2007]{veeraraghavan2007}
Veeraraghavan, H., Bird, N., Atev, S., and Papanikolopoulos, N. (2007).
\newblock Classifiers for driver activity mo\-ni\-to\-ring.
\newblock {\em Transportation Research Part C: Emerging Technologies},
  15(1):51--67.

\bibitem[Viola and Jones, 2001]{viola2001}
Viola, P. and Jones, M. (2001).
\newblock Robust real-time object detection.
\newblock {\em International Journal of Computer Vision}, 57(2):137--154.

\bibitem[Wang, 2005]{wang2005}
Wang, L. (2005).
\newblock {\em Support Vector Machines: theory and applications}, volume 177.
\newblock Springer, Berlin, Germany.

\bibitem[Watkins et~al., 2011]{watkins2011}
Watkins, M.~L., Amaya, I.~A., Keller, P.~E., Hughes, M.~A., and Beck, E.~D.
  (2011).
\newblock Autonomous detection of distracted driving by cell phone.
\newblock In {\em Intelligent Transportation Systems (ITSC), 2011 14th
  International IEEE Conference on}, pages 1960--1965. IEEE.

\bibitem[Yang et~al., 2011]{yang2011}
Yang, J., Sidhom, S., Chandrasekaran, G., Vu, T., Liu, H., Cecan, N., Chen, Y.,
  Gruteser, M., and Martin, R.~P. (2011).
\newblock Detecting driver phone use leveraging car speakers.
\newblock In {\em Proceedings of the 17th annual international conference on
  Mobile computing and networking}, pages 97--108. ACM.

\end{thebibliography}

}

\vfill
\end{document}